%% file: main.tex
\documentclass[sigconf,screen]{acmart}
\usepackage[colorinlistoftodos]{todonotes}
\usepackage{romannum} % add roman number
\usepackage[flushleft]{threeparttable}
\usepackage{multirow}
\usepackage{booktabs}
\usepackage{array}
\usepackage{subcaption}
\usepackage{comment}

\usepackage{float}
\usepackage[linesnumbered, ruled]{algorithm2e}
\usepackage{caption}
\usepackage{amsmath}
\usepackage{xcolor,cancel}

\AtBeginDocument{%
  }

\acmYear{2025}\copyrightyear{2025}
\setcopyright{cc}
\setcctype[4.0]{by}
\acmConference[MAIoT '25]{Middleware for Autonomous AIoT Systems in the Computing Continuum}{December 15--19, 2025}{Nashville, TN, USA}
\acmBooktitle{Middleware for Autonomous AIoT Systems in the Computing Continuum (MAIoT '25), December 15--19, 2025, Nashville, TN, USA}
\acmDOI{10.1145/3774901.3778066}
\acmISBN{979-8-4007-2304-9/25/12}

\begin{document}

\title{ML Inference Scheduling with Predictable Latency}


%\begin{comment}

\author{Haidong Zhao}
\orcid{0009-0009-9864-5520}
\affiliation{
  \institution{Inria \& Sorbonne University}
  \city{Paris}
  \country{France}
}
\email{haidong.zhao@inria.fr}

\author{Nikolaos Georgantas}
\orcid{0000-0001-5704-4889}
\affiliation{
  \institution{Inria}
  \city{Paris}
  \country{France}
}
\email{nikolaos.georgantas@inria.fr}

%\end{comment}

%\author{{\rm Paper \#108}}
 
%\author[1]{Florian Suri-Payer}
%\author[1]{Matthew Burke}
%\author[1]{Zheng Wang}
%\author[1]{Yunhao Zhang}
%\author[1]{Lorenzo Alvisi}
%\author[2]{Natacha Crooks}
%\affil[1]{Cornell University}
%\affil[2]{UC Berkeley}

%\author{\large {\rm Haidong Zhao}
%\\
%\vspace{2pt}\normalsize {\it Inria \& Sorbonne Universit\'{e}}}


\begin{abstract}
Machine learning (ML) inference serving systems can schedule requests to improve GPU utilization and to meet service level objectives (SLOs) or deadlines.
However, improving GPU utilization may compromise latency-sensitive scheduling, as concurrent tasks contend for GPU resources and thereby introduce interference.
Given that interference effects introduce unpredictability in scheduling, neglecting them may compromise SLO or deadline satisfaction.
Nevertheless, existing interference prediction approaches remain limited in several respects, which may restrict their usefulness for scheduling.
First, they are often coarse-grained, which ignores runtime co-location dynamics and thus restricts their accuracy in interference prediction.
Second, they tend to use a static prediction model, which may not effectively cope with different workload characteristics.
In this paper, we evaluate the potential limitations of existing interference prediction approaches, finding that coarse-grained methods can lead to noticeable deviations in prediction accuracy and that static models degrade considerably under changing workloads.
\end{abstract}







\maketitle

\input{mainmatter/Prelude}

\input{mainmatter/Development}
\input{mainmatter/Finale}

\begin{acks}

We thank the anonymous reviewers for their helpful feedback and our colleague Émile Royer for his reading.

\end{acks}

\bibliographystyle{ACM-Reference-Format}
\bibliography{bibliography/misc,bibliography/Systems,bibliography/latency}

%\appendix

\end{document}

%% file: mainmatter/Prelude.tex
\section{Introduction}
\label{maiot:introduction}

%Machine learning (ML) techniques and deep neural network (DNN) models underpin a diverse array of domains, spanning Internet applications~\cite{covington_deep_2016,google_search_engine, real_ml}, industrial scenarios~\cite{xu_high-throughput_2023,cline_predictive_2017,davari_predictive_2021,noauthor_audis_2024}, and analytics~\cite{bhardwaj_ekya_2022}.
Machine learning (ML) techniques and deep neural network (DNN) models have been transforming a diverse array of domains, spanning Internet applications~\cite{covington_deep_2016,google_search_engine, real_ml}, industrial scenarios~\cite{latency_edge_ai}, and analytics~\cite{nam_deep_2020}.
In these cases, utilizing trained DNN models to make timely predictions, or \emph{inference}, directly influences their practical impact.
For example, Amazon has reported that every 100 ms of additional latency results in a 1\% loss in sales~\cite{latency_amazon}. Similarly, untimely responses during visual inspection in manufacturing may allow defective items to pass unchecked in high-throughput pipelines~\cite{latency_edge_ai}.
Consequently, many applications specify their service level objectives (SLOs) or deadlines for their inference tasks.

ML inference workloads are often accelerated by specialized hardware such as GPUs.
Despite their effectiveness, GPUs are costly resources, both in terms of acquisition and operational overhead~\cite{gpu_cpu_1}.
Consequently, enabling multi-tenancy by deploying multiple DNN models on shared infrastructure has become a widely adopted strategy to improve cost efficiency.
To support this strategy, ML inference serving systems~\cite{olston_tensorflow-serving_2017,triton_inference_server,torch_serve} manage system resources across deployed models, aiming to utilize GPU resources efficiently while ensuring satisfied performance.

\begin{figure}[t]
  \centering
  \includegraphics[width=\linewidth]{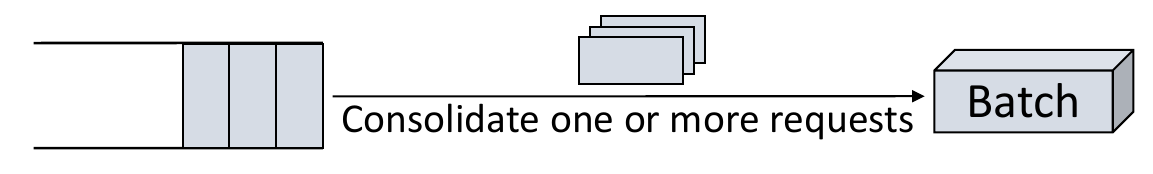}
  \caption{Requests to the same models can be consolidated and executed as a single execution unit, or batch, to improve GPU utilization.}
  \label{figure:mc_batch_formation}
\end{figure}

Achieving high GPU utilization while meeting the latency constraints of deployed models remains a challenging task.
To enhance GPU utilization, a common practice~\cite{crankshaw_clipper_2017, olston_tensorflow-serving_2017, triton_inference_server} is to employ \emph{batching}, in which multiple requests to the same model are executed simultaneously to amortize memory access overheads, as illustrated in Figure~\ref{figure:mc_batch_formation}.
A larger batch size—before saturating GPU resources—may increase the overall execution time; however, the average per-request execution time can decrease due to improved GPU utilization.
In practice, existing serving systems~\cite{choi_serving_2022,olston_tensorflow-serving_2017,triton_inference_server,torch_serve} may also support executing multiple batches concurrently on a GPU.
This is because a single batch may not efficiently utilize GPU resources, owing to runtime dynamics and workload characteristics. In addition, concurrent batch execution can alleviate head-of-line (HoL) blocking, thereby benefiting latency-sensitive scheduling. 
However, one factor may hinder this efficiency: concurrent batches contend for GPU resources, introducing non-negligible interference.
The extent of this interference depends on workload characteristics and system configurations, and it can vary considerably.
When additional runtime factors—such as queuing delays and system overheads—are considered, achieving a high level of SLO or deadline satisfaction becomes challenging without knowledge of the interference effects at scheduling time.
Otherwise, the scheduled batch may violate the latency constraints of ongoing tasks or even fail to meet its own latency constraint.

Given the complex contention among concurrent batches, interference prediction appears to be a promising approach for estimating this effect.
However, existing approaches~\cite{mendoza_interference-aware_2021, yeung_horus_2022, choi_serving_2022, kim_interference-aware_2024} for DNN workloads are limited in several aspects.
First, they are often coarse-grained, as they may ignore the fact that a batch can co-locate with different batches of varying durations.
Second, their prediction models are static, while the workload characteristics may vary over time.
In this paper, we focus on evaluating whether coarse-grained interference prediction may lead to deviations in accuracy, and on assessing whether prediction performance degrades under changing workload characteristics.

\section{Background}
\label{maiot:background}
GPUs (we mainly discuss NVIDIA GPUs in this paper) possess massive parallel resources, in which compute units are uniform and replicated.
These architectures handle large-scale parallel processing efficiently through the Single Instruction, Multiple Threads (SIMT) execution model, in which the GPU executes the same instruction across multiple threads concurrently.
Consequently, these accelerators are well-suited for DNN workloads, whose operations can be effectively parallelized and vectorized. In particular, most DNN computations—such as matrix multiplications and convolutions—exhibit high arithmetic intensity and data parallelism, enabling efficient mapping onto the GPU's massively parallel and high-throughput architecture.
To effectively exploit GPU resources, we discuss common practices for improving GPU utilization in inference workloads and analyze the potential interference effects that arise from resource contention.

\begin{figure}[t]
  \centering
  \includegraphics[width=\linewidth]{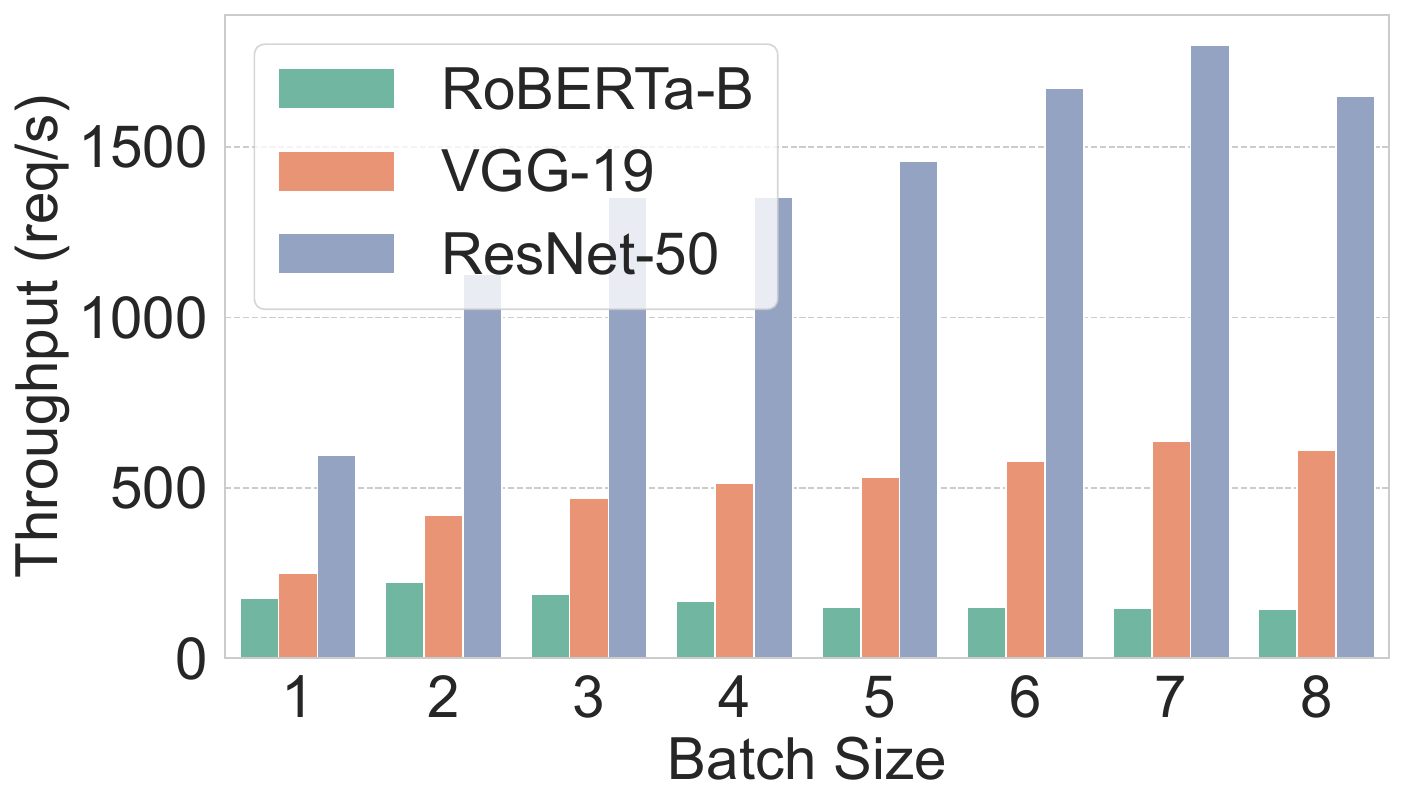}
  \caption{Request throughput across varying batch sizes for each model.
  This benchmark calculates throughput solely based on the profiled inference execution duration, without accounting for runtime dynamics or data transfer latency between the host and GPU memory. 
  Therefore, the reported values represent a theoretical maximum throughput.}
  \label{figure:maiot_batch_throughput}
\end{figure}

\textbf{Dynamic batching.}
Figure~\ref{figure:maiot_batch_throughput} presents the theoretical maximum throughput across different batch sizes for each model.
Larger batch sizes often improve request throughput, but the magnitude of improvement depends on the model characteristics and available resources.
Lightweight models such as ResNet-50~\cite{he_deep_2016} can achieve more than a 200\% increase when the batch size is enlarged; in contrast, RoBERTa-B~\cite{liu_roberta_2019} gains only 26.5\%.
This suggests that once a heavy model saturates GPU resources or reaches its bottleneck, increasing the batch size further is unlikely to improve, or even hurt throughput.
The dynamic batching technique can be employed to exploit the benefits of batching. It allows developers to configure a waiting time for a model. Consequently, after the first request for that model arrives in the request queue, it may momentarily wait for subsequent requests. This allows the system to aggregate them into a larger batch.
However, a larger batch size typically incurs higher inference latency. Since requests still need to meet their SLOs or deadlines, the batch size must be carefully tuned when making scheduling decisions.

\begin{figure}[t]
  \centering
  \includegraphics[width=\linewidth]{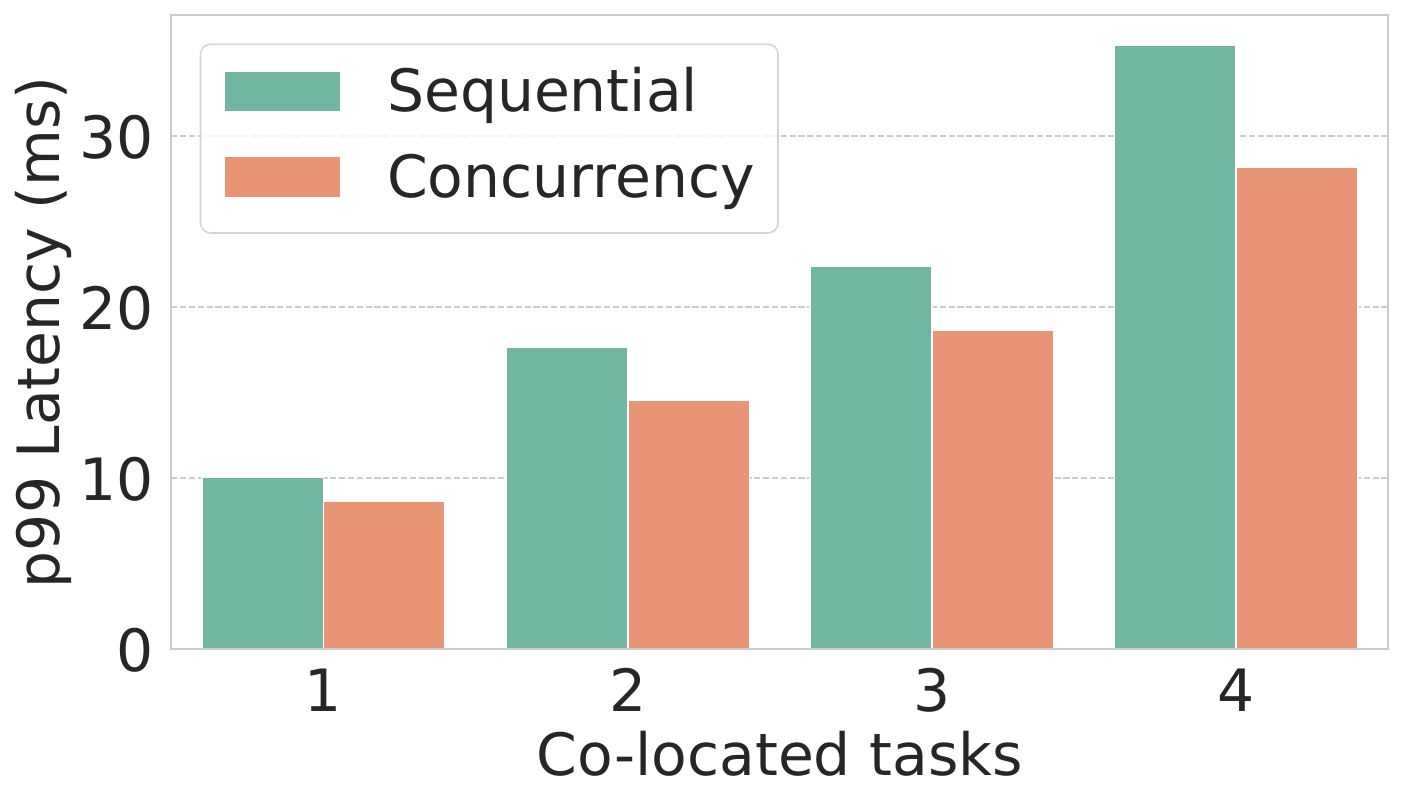}
  \caption{
  The number of co-located tasks on a GPU with respect to the 99th percentile latency under sequential execution and concurrent execution, respectively.
  All tasks deploy the same model, ResNet-50~\cite{he_deep_2016}, with inference loads evenly distributed among them. 
  Under stress testing, larger batch sizes may readily aggregate and thereby saturate GPU resources; however, employing concurrency can still help mitigate HoL blocking.}
  \label{figure:maiot_concurrency}
\end{figure}

\textbf{Concurrent batch execution.}
Executing multiple batches concurrently can further improve GPU utilization. This strategy is particularly beneficial when a single batch does not fully saturate GPU resources, or when some batches incur noticeable data transfer overhead between host and GPU memory.
Furthermore, employing concurrency may mitigate HoL blocking. 
Figure~\ref{figure:maiot_concurrency} shows that, under stress testing, the concurrency-enabled approach achieved lower 99th-percentile latency across all co-located scenarios. This suggests that concurrency may lead to more balanced performance among deployed models, and in turn, support latency-sensitive scheduling.

\begin{figure*}[t]
  \centering

  \begin{subfigure}{0.32\linewidth}
    \centering
    \includegraphics[width=\linewidth]{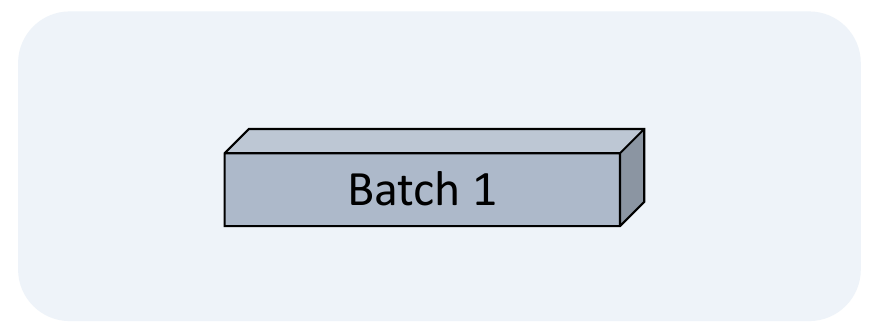}
    \caption{At this moment, only Batch 1 is running in the system.}
    \label{fig:mc_timeline_1}
  \end{subfigure}\hfill
  \begin{subfigure}{0.32\linewidth}
    \centering
    \includegraphics[width=\linewidth]{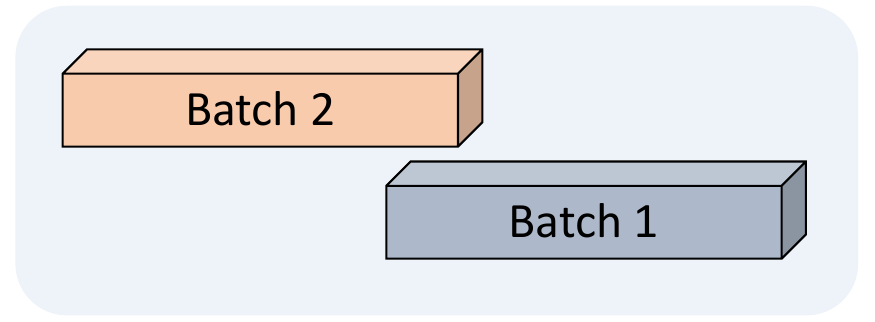}
    \caption{Batch 2 arrives and co-locates briefly with Batch 1 until its departure.}
    \label{fig:mc_timeline_2}
  \end{subfigure}\hfill
  \begin{subfigure}{0.32\linewidth}
    \centering
    \includegraphics[width=\linewidth]{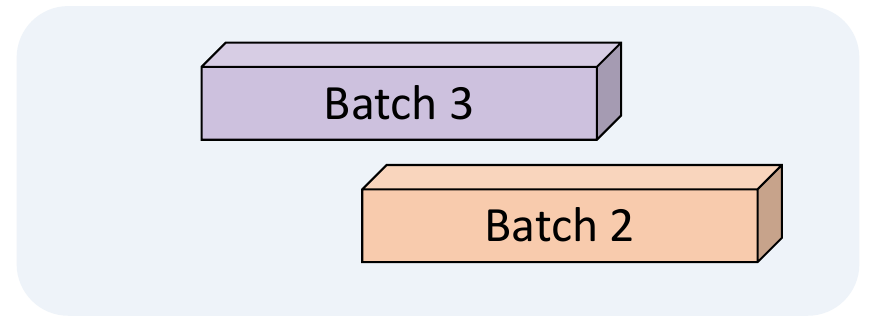}
    \caption{Batch 3 arrives and co-locates for a relatively long period with Batch 2.}
    \label{fig:mc_timeline_3}
  \end{subfigure}

  \caption{Illustration of potential batch co-location dynamics in chronological order.}
  \label{fig:mc_timeline}
\end{figure*}

\textbf{Kernel execution interference.}
Executing batches concurrently may improve GPU utilization and support more balanced performance; however, it can also introduce interference, since the batches compete for limited GPU resources.
On GPUs, each batch is executed as a sequence of compiled kernels, which are functions designed to exploit the accelerator's parallel execution capabilities.
When batches execute concurrently, their respective kernels contend for GPU memory resources (e.g., L2 cache, DRAM) and compute resources (e.g., Tensor Cores).
The resulting interference effects could be difficult to model.
First, concurrent kernels may have heterogeneous resource demands (e.g., compute-bound or memory-bound), which leads to varying degrees of interference~\cite {strati_orion_2024}. 
Second, since each batch is composed of a sequence of kernels, it may be infeasible to determine the exact temporal overlap of kernels within two or multiple batches.
Accordingly, a tractable approach may rely on predictions made at the batch granularity.

%% file: mainmatter/Development.tex
\section{Challenge}
\label{maiot:challenge}

The interference-induced delays introduced unpredictability, thereby complicating latency-sensitive scheduling.
Through simulations of heterogeneous workloads under various setups (including stress testing), where each workload comprises a set of deployed models with varying request arrival rates, noticeable interference can be introduced. 
Specifically, the interference—defined as the ratio of the measured kernel execution time to its profiled value (95th-percentile)—can reach up to 151.66\% at the 95th-percentile when the maximum permitted batch concurrency is set to 2.
The results suggest that even if a batch experiences resource contention from only one concurrent batch at a given time, severe interference can still arise. Furthermore, the magnitude of interference may escalate as the concurrency level increases.
Consequently, we suppose that predicting kernel execution interference is a prerequisite for achieving high GPU utilization while maintaining a high degree of SLO or deadline satisfaction.
However, existing approaches~\cite{kim_interference-aware_2024,mendoza_interference-aware_2021,choi_serving_2022,yeung_horus_2022,kim_interference-aware_2021} for DNN workloads are limited in several aspects.

\subsection{Coarse-Grained}

Even if inference workloads operate at the millisecond level, a single batch may still encounter multiple co-location scenarios during execution. Figure~\ref{fig:mc_timeline} illustrates the potential co-location dynamics on a GPU in chronological order.
Figure~\ref{fig:mc_timeline_1}: Initially, only Batch 1 is executed, and thus no interference occurs.
Figure~\ref{fig:mc_timeline_2}: Subsequently, Batch 2 arrives and co-locates with Batch 1 for a relatively short duration, until Batch 1 departs.
Figure~\ref{fig:mc_timeline_3}: Later, Batch 3 arrives and co-locates with Batch 2 for a relatively long duration, until Batch 2 departs.
Consequently, the co-location dynamics differ for each batch:
Batch 1 has a short co-location duration with Batch 2;
Batch 2 has a short co-location duration with Batch 1 and a long co-location duration with Batch 3;
Batch 3 has a long co-location duration with Batch 2.

However, existing approaches~\cite{kim_interference-aware_2024,mendoza_interference-aware_2021,choi_serving_2022,yeung_horus_2022,kim_interference-aware_2021} often overlook these dynamics, thereby limiting their ability to estimate interference.
They typically account for interference effects only for the batch currently being scheduled, while ignoring subsequent co-location dynamics, such as when co-located batches depart or new batches arrive.
For example, when scheduling Batch 2, they may consider only the current runtime dynamics: Batch 1 is running, and therefore assume that interference effects for Batch 2 stem from Batch 1. Consequently, accurate interference prediction can potentially be achieved only if Batch 2 co-locates with Batch 1 for most of its execution time.

In the presented scenarios, Batch 1 and Batch 2 co-locate only briefly. Batch 2 subsequently undergoes a relatively prolonged co-location with Batch 3, where the interference effects are more likely attributable to the latter.
If Batch 1 differs markedly in resource demands from Batch 3, noticeable prediction errors may be inevitable.
Accordingly, we posit that neglecting co-location dynamics and the corresponding temporal effects may lead to noticeable deviations in interference prediction.

\subsection{Non-Adaptive}

Existing approaches~\cite{kim_interference-aware_2024,mendoza_interference-aware_2021,choi_serving_2022,yeung_horus_2022,kim_interference-aware_2021} train prediction models offline and subsequently use them for scheduling. 
However, the challenge of handling potential data and concept drift (e.g., newly deployed models, changing load patterns) has received little attention.
This suggests that if accuracy degrades under changing workload characteristics, developers may need to retrain the model using more recent serving data and redeploy it.

To mitigate this issue, online learning can be used to update prediction models dynamically during runtime. By continuously adapting to recent serving data, it allows the prediction model to respond to changes in workload characteristics and potentially sustain predictive accuracy at an acceptable level.

\section{Evaluation}
\label{maiot:evaluation}
We attempt to answer the following questions:
(\romannum{1}) What is the prediction accuracy if co-location dynamics are ignored?
(\romannum{2}) Can prediction accuracy be improved if the model is adaptive?

\subsection{Methodology}
\label{sec:maiot_methodology}

\noindent
\textbf{Experimental Setup.}
The experiment is conducted on an NVIDIA L4 GPU with 24 GB of DRAM within a virtualized cloud instance.
TensorRT~\cite{tensorrt_doc} is employed as the inference engine to optimize model execution.

\noindent
\textbf{Workloads.}
We evaluate DNN models with CNN and Transformer architectures, including ResNet-50~\cite{he_deep_2016}, YOLO-v8n~\cite{jocher_ultralytics_2023}, RoBERTa-B~\cite{liu_roberta_2019}, ViT-B-16~\cite{dosovitskiy_image_2021}, VGG-19~\cite{simonyan_very_2015}, and ConvNeXt-B~\cite{liu_convnet_2022}. 
The maximum batch size is set to 8 for each model.
The request arrival rate follows a Poisson process.

\noindent
\textbf{Interference Prediction Approach.} 
We adopt the method from \emph{gpulets}~\cite{choi_serving_2022}, which uses a linear regression model:
\[
\hat{y} = w^{\top} x + b,
\]
where \(x\) denotes the input feature vector, \(w\) is the learned weight vector, and \(b\) is the bias term, jointly determining the prediction \(\hat{y}\).

Specifically, it uses the average resource throughput of the L2 cache and DRAM metrics of the interfered batch and its co-located batch, profiled offline with NVIDIA Nsight Compute~\cite{nvidia_nsight_compute}, as input features.
Since \emph{gpulets} employs Multi-Process Service (MPS)~\cite{nvidia_mps} to partition GPU compute resources among the deployed models, thereby eliminating interference in compute resources.
To capture (overview) compute activity, we further include the \textit{Streaming Multiprocessors (SMs) Throughput} metric from Nsight Compute~\cite{nvidia_nsight_compute} as part of the input features.
To more faithfully reproduce the evaluated environment in \emph{gpulets}, we limit the maximum batch concurrency per GPU to 2, ensuring that no more than two batches run concurrently at any given time.

\subsection{Coarse-Grained Interference Prediction}
As discussed, each batch (a model and batch-size pair) is profiled offline to obtain its average resource throughput.
The resource throughput of the co-located batch serves as the input features for the interference prediction model.
However, the set of co-located batches may change during a batch’s execution, and consequently, its co-located resource throughput may also vary.
We examine two approaches to calculating the co-located resource throughput: one that fully ignores co-location dynamics, and another that partially accounts for co-location dynamics while overlooking temporal effects, i.e., disregarding the co-location duration of any batch.

\noindent
(1) Fully ignoring co-location dynamics: when a batch is scheduled, its co-located resource throughput is determined by the current runtime dynamics (the resource throughput of ongoing batches) and remains unchanged thereafter.

\noindent
(2) Partially considering co-location dynamics: For each batch, we track the arrivals and departures of its co-located batches, along with the corresponding changes in co-located resource throughput. Consequently, a batch may experience fluctuating co-located resource throughput over its execution. To obtain an overall estimate, we apply an Exponentially Weighted Moving Average (EWMA) to smooth these potentially fluctuating values.

\[
\hat{R}_t = \alpha \cdot X_t + (1-\alpha) \cdot \hat{R}_{t-1}, \; \alpha \in (0,1]
\]

Here, $\hat{R}_t$ is the EWMA-smoothed co-located resource throughput, $X_t$ denotes the observed co-located resource throughput at time $t$, and $\alpha$ is the smoothing factor that controls the relative weight of the most recent observation with respect to the past estimate. we consider smoothing factors $\alpha \in \left\{\tfrac{1}{3}, \tfrac{1}{2}, \tfrac{2}{3}\right\}$ in the following evaluation.

\begin{figure}[t]
  \centering
  \includegraphics[width=\linewidth]{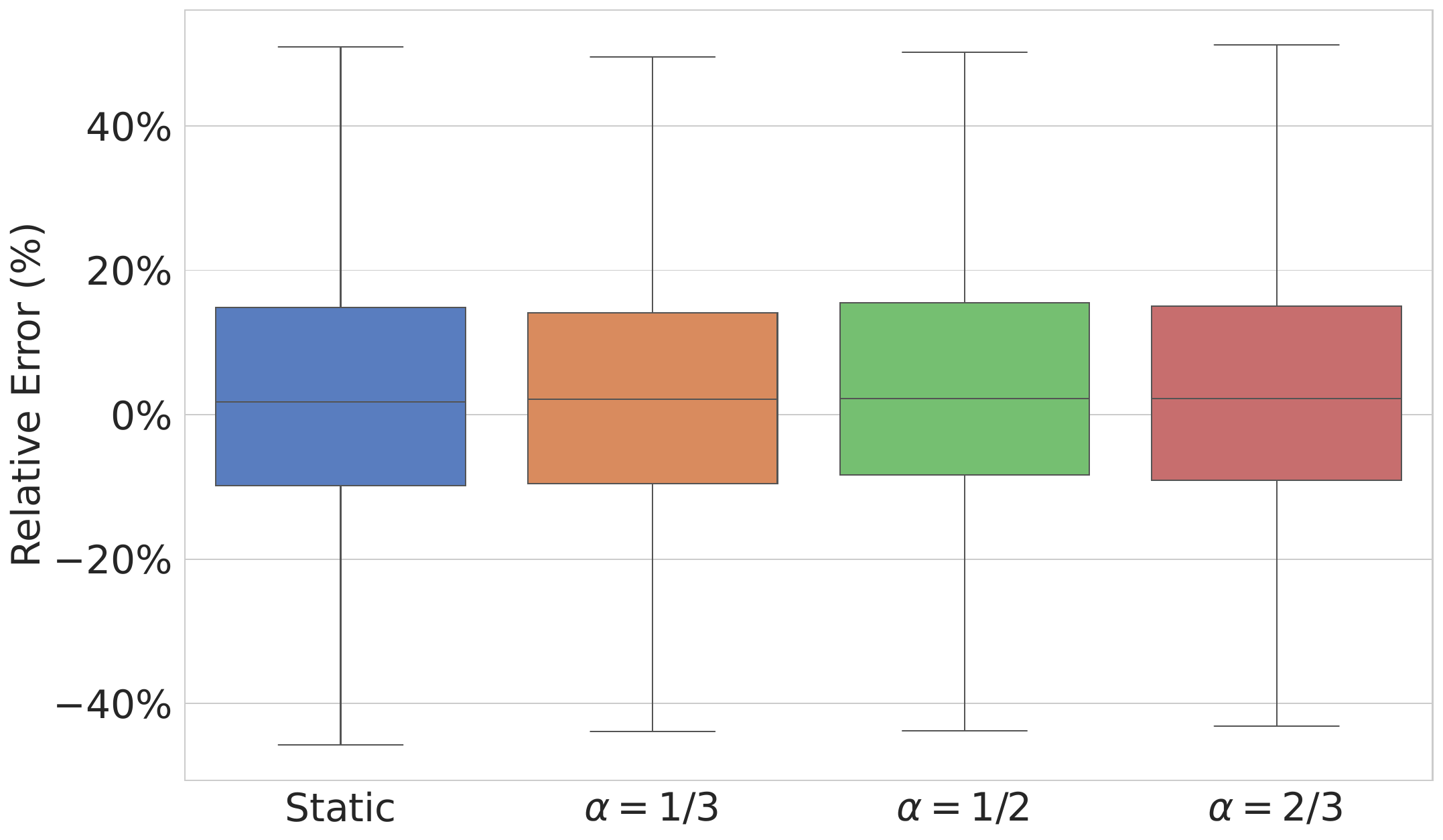}
  \caption{Relative interference prediction error when runtime dynamics are fully ignored (Static) or partially considered (EWMA under varying $\alpha$).}
  \label{figure:mc_intf_calculation}
\end{figure}

We simulate a set of workloads, with each case varying in at least one aspect: the deployed models, the request arrival rates, or the load intensities, which range from light to stress-testing levels.
During these simulations, the serving data, including the interference prediction results, is logged and partitioned into a training set (75\%) and a test set (25\%). The prediction model is then trained on the training set and evaluated on the test set.
Figure~\ref{figure:mc_intf_calculation} shows that the interference prediction achieves comparable performance across all cases.
More than 50\% of the predictions lie within relative errors below 20\%, as indicated by the interquartile range. However, noticeable deviations occur, with some predictions exhibiting errors exceeding 60\%.
We hypothesize that this high deviation arises from using inaccurate co-located resource throughput values as inputs during both training and testing.
In training, such inaccurate input features limit the development of a more accurate prediction model. In testing, they fail to capture the actual resource pressure experienced by a batch, which in turn reduces prediction quality.
This indicates that whether co-location dynamics are fully ignored or only partially considered, noticeable deviations may arise in the interference prediction.

\subsection{Static Prediction Model}

To evaluate whether the static interference prediction model suffers from decreased accuracy under changing workload characteristics, we collect several serving datasets under different conditions. In each condition, a set of DNN models is deployed in the serving system, and the corresponding interference prediction data are logged.

\begin{itemize}
\item Training Set: We deploy a partial set of evaluated models into the serving system.

\item Test Set 1: Compared with the \emph{Training Set}, we adjust only the loads (i.e., the request arrival rates) for each deployed model, which makes them differ markedly.

\item Test Set 2: We select an alternative set of deployed models, none of which are deployed in \emph{Training Set}.

\item Test Set 3: This set comprises all the deployed models from the above sets, thus representing heterogeneous workloads.
\end{itemize}

We compare three methods: \emph{offline learning}, which only trains the prediction model on the \emph{Training Set}, representing the case where a static model is deployed in real-world settings; and \emph{online learning}, which updates the prediction model at runtime based on prediction errors. 
In the online learning setting, we compare parameters that are updated either through plain stochastic gradient descent (SGD) or through recursive least squares (RLS). 
All three methods are initially fitted on the \emph{Training Set}, which ensures identical starting parameters and mean squared errors (MSE).

\begin{table}[t]
\centering
\caption{Comparison of prediction error (MSE) among offline learning, online learning with the SGD optimizer, and online learning with RLS.}
\label{tab:maiot_mse_compare}
\begin{tabular}{lccc}
\toprule
 & \multicolumn{3}{c}{\textbf{Method}} \\
\cmidrule(lr){2-4}
\textbf{Dataset} & \textbf{Offline} & \textbf{SGD} & \textbf{RLS} \\
\midrule
Training Set & 0.0355 & 0.0355 & 0.0355 \\
Test Set 1   & 0.0394 & 0.0346 & 0.0343 \\
Test Set 2   & 0.5480 & 0.4311 & 0.3467 \\
Test Set 3   & 0.2065 & 0.1679 & 0.1499 \\
\bottomrule
\end{tabular}
\end{table}

Table~\ref{tab:maiot_mse_compare} reports the resulting MSE for all methods on the training and test sets, respectively.
For load variance (Test Set 1), the offline learning method shows a slight increase in prediction error, whereas the two online learning methods maintain comparable error levels. When coping with interference data from previously unseen deployed models (Test Set 2), the prediction error rises significantly across all methods; however, the online learning methods yield relatively lower errors. 
In the heterogeneous workloads (Test Set 3), which reflects a heterogeneous workload, online learning methods again outperform the offline learning method.
Interestingly, all methods perform better in \emph{Test Set 3} than in \emph{Test Set 2}. We hypothesize that this improvement arises because \emph{Test Set 3} contains the deployed model used during initial training. As a result, the interference prediction data in the \emph{Training Set} has already shaped the model’s weights, bringing them closer to the characteristics of that deployed model. 
In contrast, the model weights are largely unfamiliar with the characteristics of the deployed models in \emph{Test Set 2}, which leads to noticeably poorer prediction accuracy.

These results suggest that incorporating online learning into prediction models may help cope with changing workload characteristics. 
Moreover, RLS consistently outperforms plain SGD, likely because RLS exploits curvature information to achieve faster convergence, albeit at a higher computational cost. This indicates that the choice of optimization algorithms can affect predictive performance.

%% file: mainmatter/Finale.tex
\section{Related Work}
\label{maiot:related_work}

Interference prediction has been employed in many scenarios, ranging from modeling resource contention of co-located applications in multi-core servers~\cite{xu_pythia_2018}, to mitigating I/O latencies in storage systems~\cite{hao_linnos_2020}, and to enabling informed scheduling decisions for DNN workloads~\cite{kim_interference-aware_2024,mendoza_interference-aware_2021,choi_serving_2022, yeung_horus_2022, kim_interference-aware_2021}.
This paper discusses DNN workloads, which generally consist of two phases: \emph{training} and \emph{inference}.
Inference workloads are closely related to their corresponding training workloads. In each iteration, training performs a forward pass to compute intermediate values and a backward pass to update the model weights using the gradients. Inference, in contrast, executes only the forward pass without any subsequent gradient computation or weight update.
Nevertheless, their characteristics can still differ markedly within the forward pass.
In particular, training workloads typically use a relatively large mini-batch (e.g., 32 or 64 samples) to ensure stable convergence.
This contrasts with inference workloads, where the chosen batch size may be subject to latency constraints, runtime dynamics, and configurations.
Consequently, some metrics used as a proxy for estimating interference in training workloads~\cite{yeung_horus_2022,kim_interference-aware_2021} may not apply to inference workloads. For example, \textit{GPU Utilization}, as reported by \texttt{nvidia-smi}, reflects the percentage of time that any operation runs on the GPU, regardless of its intensity~\cite{nvidia_gpu_utilization}.

Xu et al.~\cite{xu_characterization_2019} study interference for DNN workloads in NVIDIA vGPU, which shares GPUs among virtual machines (VMs). 
They consider several resource metrics and accordingly select a candidate model that achieves the best training results.
However, vGPU technology time-slices GPUs across VMs rather than supporting true concurrency, thereby introducing non-negligible context-switching overheads. This makes the vGPU technology potentially unsuitable for inference serving.

Mendoza et al.~\cite{mendoza_interference-aware_2021} leverage both runtime resource metrics and machine-type embeddings (learned vectors that place similar hardware types close together, allowing the predictor to generalize to under-profiled machines) to enhance interference prediction.
Their approach, however, does not exploit batching. Moreover, it operates at the cluster level, which can introduce non-negligible overheads and complicate scaling to nodes with heterogeneous GPU types and varying numbers of GPUs.

Choi et al.~\cite{choi_serving_2022} propose an inference server that leverages MPS~\cite{nvidia_mps} to partition GPU computing resources and employs a linear regression model to estimate interference caused by resource contention in the GPU memory hierarchy. We adopt this approach in our evaluation, and the results indicate that it may struggle to deliver satisfactory prediction performance.

Kim et al.~\cite{kim_interference-aware_2024} propose an interference prediction model for NVIDIA Jetson, which is orthogonal to our focus on discrete GPU architectures. 
Jetson is a System-on-Chip (SoC) that integrates heterogeneous processing subsystems, including a CPU, a GPU, and two dedicated Deep Learning Accelerators (DLAs).
Due to this heterogeneity, multiple prediction models must be maintained independently; for example, one model for tasks running on the GPU, which may encounter resource contention from one or both DLAs, and, conversely, models for the DLAs that may experience contention from the GPU.

\section{Work in Progress}
\label{maiot:conclusion}

We have discussed the potential limitations of existing interference prediction approaches for ML inference workloads.
Notably, we argued that ignoring runtime co-location dynamics may lead to considerable deviations in predictions, and the accuracy of prediction models may degrade if workload characteristics change.
Consequently, we suppose that interference prediction for inference workloads should be reexamined and redesigned to account for these factors.
We outline our ongoing work to achieve efficient ML inference scheduling as follows.

\textbf{Generalize to cloud and on-premises scenarios.}
ML inference serving systems serve as supporting infrastructures for both cloud and on-premises environments. 
We aim to improve interference prediction methods and integrate them into the serving system to enhance SLO or deadline satisfaction.
We begin with discrete GPUs, because they are more commonly deployed in both cloud and on-premises settings.

\textbf{Improve prediction model.} 
We have evaluated the interference prediction approach used in \emph{gpulets}~\cite{choi_serving_2022}. 
To improve interference prediction, we proceed in the following directions.
1) Selecting more representative resource metrics (e.g., L2 Cache throughput) that better capture the non-linear behavior of the GPUs. Meanwhile, extend the support to cases where more than two batches run concurrently, since real systems often host tens of models with higher concurrency levels and mixed operators.
2) Constructing more sophisticated prediction models, while carefully balancing the trade-off between accuracy and prediction latency.
3) Exploring optimal optimization algorithms or optimizers for gradient descent for online learning.